\documentclass[10pt,twoside, submission]{article}

\usepackage{times}
\usepackage[utf8]{inputenc}
\usepackage[T1]{fontenc}
\usepackage{graphicx}

\usepackage{algorithm}
\usepackage{algorithmic}
\usepackage{amsmath}
\usepackage[dvipsnames]{xcolor}

\usepackage{rotating}

\definecolor{commentcolor}{gray}{0.3}

\usepackage{taln2019}
\usepackage[frenchb]{babel}

\title{Apprentissage de plongements lexicaux par une approche réseaux complexes}

\author{Victor Connes\up{1, 2}\quad Nicolas Dugué\up{2}\\
 {\small
    (1) Le Mans Université, LIUM, EA 4023,
Laboratoire d’Informatique de l’Université du Mans \\ 
    (2) LS2N Université de Nantes – faculté des Sciences et Techniques (FST) Bâtiment 34 2 Chemin de la Houssinière BP 92208, 44322 Nantes Cedex 3 \\ 
    \texttt{
      victor.connes@univ-nantes.fr, nicolas.dugue@univ-lemans.fr \\ 
}}}

\begin{document}
\maketitle

\resume{
    La littérature des réseaux complexes a montré la pertinence de l'étude de la langue sous forme de réseau pour différentes applications : désambiguïsation, résumé automatique, classification des langues, \textit{etc.} 
    Cette même littérature a démontré que les réseaux de co-occurrences de mots possèdent une structure de communautés latente. 
    Nous formulons l'hypothèse que cette structuration du réseau sous forme de communautés est utile pour travailler sur la sémantique d'une langue et introduisons donc dans cet article une méthode d'apprentissage de plongements originale basée sur cette hypothèse. Cette hypothèse est cohérente avec la proximité qui existe entre la détection de communautés sur un réseau de co-occurrences et la factorisation d'une matrice de co-occurrences, méthode couramment utilisée pour l'apprentissage de plongements lexicaux. Nous décrivons notre méthode structurée en trois étapes : construction et pré-traitement du réseau, détection de la structure de communautés, construction des plongements de mots à partir de cette structure. 
     Après avoir décrit cette nouvelle méthodologie, nous montrons la pertinence de notre approche avec des premiers résultats d'évaluation sur les tâches de catégorisation et de similarité. Enfin, nous discutons des perspectives importantes d'un tel modèle issu des réseaux complexes : les dimensions du modèle (les communautés) semblent interprétables, l'apprentissage est rapide, la construction d'un nouveau plongement est presque instantanée, et il est envisageable d'en expérimenter une version incrémentale pour travailler sur des corpus textuels temporels.
}

\abstract{Complex networks based word embeddings.}{
    Most of the time, the first step to learn word embeddings is to build a word co-occurrence matrix. As such matrices are equivalent to graphs, complex networks theory can naturally be used to deal with such data. In this paper, we consider applying community detection, a main tool of this field, to the co-occurrence matrix corresponding to a huge corpus. Community structure is used as a way to reduce the dimensionality of the initial space. Using this community structure, we propose a method to extract word embeddings that are comparable to the state-of-the-art approaches.
}

\motsClefs
  {Plongements lexicaux, réseaux complexes, détection de communautés}
  {Word embeddings, complex networks, community detection}

\section{Introduction}
Dans l'état de l'art de l'apprentissage de plongements lexicaux, on recense de nombreuses approches basées sur une matrice de co-occurrences termes-termes construite en utilisant de grands corpus~\cite{pennington_glove:_2014,goldberg_improving_nodate}. Les auteurs factorisent ensuite cette matrice creuse de façon à obtenir un nouvel espace dans lequel chaque terme est représenté par un vecteur dense. 

Dans le domaine des réseaux complexes, ces matrices de co-occurrences sont appelées \emph{graphes} ou \emph{réseaux}. L'étude du langage naturel par le prisme des réseaux complexes n'est pas une science nouvelle. L'état de l'art du domaine utilise également de grands corpus pour construire des réseaux $G=(V,E)$ tels que chaque n\oe{}ud $u \in V$ du réseau représente un terme du vocabulaire, et un lien $(u,v) \in E$ entre deux n\oe{}uds représente une co-occurrence dans le corpus entre deux termes. Ces réseaux peuvent être dirigés, ou valués, on se dote alors d'une fonction $w$ qui associe un poids à chaque lien $w:E \to {\rm I\!R}$. 

Ces travaux ont notamment permis de révéler plusieurs propriétés de ces réseaux et ainsi de mieux comprendre la façon dont est construite la langue : ces réseaux sont petit-monde~\cite{i_cancho_small_2001}, sans-échelle avec une loi de puissance à deux vitesses~\cite{i_cancho_small_2001} expliquée par le modèle de~\citealt{dorogovtsev_language_2001}, et le poids des liens suit également une loi de puissance dans le cas des réseaux valués~\cite{gao_comparison_2014,masucci_network_2006}.

Parmi les propriétés observées, ce papier se concentre sur la présence d'une structure de communautés dans ces réseaux~\cite{newman_analysis_2004}. La structure de communautés d'un réseau est une partition des n\oe{}uds du réseau telle que pour chaque partie, les n\oe{}uds sont plus connectés entre eux qu'avec le reste du réseau~\cite{newman2004finding}. Nous faisons l'hypothèse que cette structure de communautés permet de construire des plongements lexicaux. 

Cette hypothèse se base sur deux constats. Le premier vient des exemples de~\citealt{palla_uncovering_2005} qui semblent indiquer que les communautés encapsulent une partie de l'information sémantique. D'ailleurs, la définition de la structure de communautés vient appuyer ce constat : pour chaque partie (communauté) de la partition (structure de communautés), les n\oe{}uds sont plus connectés entre eux qu'avec le reste du réseau. Au regard de l'hypothèse de Firth "\emph{a word is characterized by the company it keeps}", on comprend que chaque communauté sera constituée de mots qui seront utiles pour se caractériser les uns les autres. Le second constat vient de certains travaux de la littérature qui mettent en évidence les liens entre décomposition en valeur singulière (SVD) et détection de communautés~\cite{sarkar_community_2011}. Or, appliquer une SVD à une matrice de co-occurrences pondérée par la \emph{positive pointwise mutual information} est une méthode efficace pour aboutir à des plongements lexicaux~\cite{goldberg_improving_nodate}.

Nous présentons donc Section~\ref{sec:methode} notre approche basée sur la détection de communautés pour extraire des plongements. Cette approche considère chaque communauté comme une dimension, et les liens d'un n\oe{}ud vers ces communautés permettent de calculer pour chaque dimension la valeur de la composante. Nous montrons Section~\ref{sec:resultats} que les résultats expérimentaux démontrent la pertinence de l'approche, d'un point de vue qualitatif, mais également quantitatif. Enfin, nous discuterons Section~\ref{sec:discussion} des avantages d'une telle approche. Tout d'abord, celle-ci permet d'espérer des dimensions interprétables. Ensuite, le calcul d'un plongement pour un terme est très rapide. Enfin, ce type d'approche ouvre des perspectives pour créer des plongements lexicaux évoluant dans le temps via des algorithmes de détection de communautés incrémentaux.

\section{Méthode}
\label{sec:methode}
\paragraph{Données.}
Les données utilisées sont les GoogleBooksNgram \footnote{http://storage.googleapis.com/books/ngrams/books/datasetsv2.html} anglais,  corpus BristishEnglish et EnglishFiction.
Les GoogleBooksNgram sont des recueils de co-occurrences de termes observées sur une grande bibliothèque de textes allant des années 1800 à 2008. Les co-occurrences sont fournies avec une fenêtre de contexte allant de 2 à 5.
Pour nos expériences, nous avons conservé seulement les co-occurrences observées depuis 1980 aboutissant à un vocabulaire avant pré-traitements d'environ $380000$ termes.

\paragraph{Construction et pré-traitement du réseau.}
Une fois le réseau créé en exploitant les co-occurrences d'un corpus textuel avec une fenêtre de taille $f$, on obtient alors $G=(V,E,w)$ comme décrit. Pour rappel, l'ensemble des n\oe{}uds $V$ est équivalent au \emph{vocabulaire} considéré, l'ensemble des liens $E$ représente les co-occurrences entre les termes du vocabulaire, et on définit la pondération des liens de $E$ avec la fonction $w(u, v)$, qui vaut le nombre de co-occurrences observées entre les termes représentés par les n\oe{}uds $u$ et $v$ dans le corpus en considérant le paramètre $f$.

Dans le but de ne conserver que les co-occurrences ayant une valeur sémantique, nous supprimons les liens entre les n\oe{}uds qui ne révèlent pas une dépendance statistique significative en utilisant l'Éq.~\ref{eq:ppmi} :
\begin{equation}
    \label{eq:ppmi}
    ppmi(w, c) = max\Bigg(0, log_2\bigg(\frac{p(u, v)}{p(v)p(u)}\bigg)\Bigg)
\end{equation}

Ce pré-traitement du réseau découle directement de ce qui est préconisé par l'état de l'art, notamment par~\citealt{goldberg_improving_nodate}. Mais il semble également pertinent de l'appliquer pour simplifier le travail de l'algorithme de détection de communautés, dont les résultats s'améliorent avec des pré-traitements de type seuillage ou repondération~\cite{yan2018weight}.
    
Dans le but d'alléger le réseau avec un filtre basse-fréquence, nous appliquons l'algorithme~\ref{alg:kcoeur} qui permet d'obtenir le $k$-c\oe{}ur du réseau~\cite{matula_smallest-last_1983} : il s'agit de supprimer tous les n\oe{}uds ayant moins de $k$ voisins de manière récursive jusqu'à que tous les n\oe{}uds restant dans le réseau soient connectés à au moins $k$ voisins.

\begin{algorithm}
    \caption{\label{alg:kcoeur} Extraction du K-c\oe{}ur}
    \begin{algorithmic}
    \label{algo:kcore}
        \REQUIRE $G=(V,E)$ graphe, $k$ entier\\
        \STATE $convergence \leftarrow False$
        \WHILE{$convergence$}
            \STATE $convergence \leftarrow True$;  $V' \leftarrow \{\}$
            \FOR{$\forall n \in V$}
                \IF{$degres(n) < k$}\STATE $V' \leftarrow V' \cup \{n\}$; $convergence \leftarrow False$ \ENDIF
            \ENDFOR
            \STATE $V \leftarrow V \setminus V'$
        \ENDWHILE
\RETURN G
    \end{algorithmic}
\end{algorithm}

Enfin, dans le but de supprimer les mots vides et de limiter l'influence des hautes fréquences (comme dans Glove~\cite{pennington_glove:_2014} ou Word2vec~\cite{mikolov2013distributed}, nous choisissons de supprimer les $ntop$ n\oe{}uds de plus haut degré. Les meilleurs résultats sont empiriquement obtenus pour $ntop$ = 200 et $k$ = 10. Après pré-traitement, nous aboutissons à un vocabulaire de $135.000$ mots dans le cas de notre corpus.

\paragraph{Détection de communautés.}

Une fois le réseau généré et pré-traité, la seconde étape consiste à détecter les communautés qui serviront par la suite de dimensions aux vecteurs de plongement lexicaux. On dit que $C$ est une partition de $V$ telle que $C = \{C_0, C_1, ..., C_{n}\}$ avec $\cup_i C_i \in C = V$. S'il n'existe pas de définition unique du concept de communauté, une structure de communautés est souvent définie comme une partition du réseau telle que les nœuds de chaque partie sont plus connectés entre eux qu'avec le reste du réseau.

\begin{figure}
    \centering
    \includegraphics[width=0.8\textwidth]{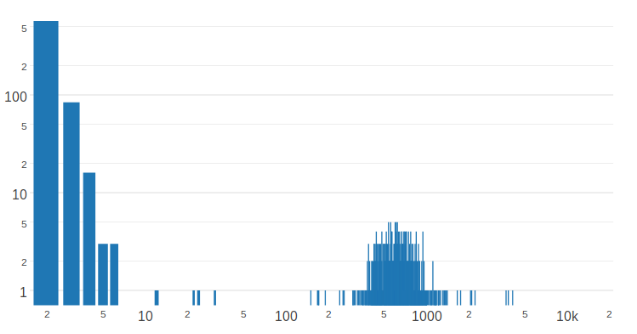}
    \caption{Distribution de la taille des communautés (en log-log) \label{fig:distribution}}
\end{figure}

De nombreuses méthodes existent pour réaliser l'extraction de ces communautés. Nous avons choisi l'algorithme~\ref{algo:LP} de propagation de labels introduit par~\citealt{raghavan_near_2007}, 
dont la complexité est quasi-linéaire en $O(|V|)$, et qui génère \emph{théoriquement} des communautés dont les tailles suivent une distribution permettant d'éviter d'avoir en grand nombre des communautés trop grandes (fourre-tout) ou trop petites (trop spécifiques)~\cite{dao_estimating_2018}. En pratique, on constate Figure~\ref{fig:distribution} un très grand nombre de petites communautés. Il s'agirait de considérer des adaptations de l'algorithme de propagation de labels pour éviter cet écueil.

\begin{algorithm}
    \caption{Propagation de labels}
    \begin{algorithmic}
    \label{algo:LP}
        \REQUIRE $G=(V, E, w)$ graphe\\
        \STATE $\forall n \in V, c(n) \leftarrow n$\\
        \STATE \textcolor{commentcolor}{\textit{On parle de convergence lorsque la communauté de chaque n\oe{}ud est la communauté majoritaire de ses voisins.}}\\
        \WHILE{$check\_convergence(G, C, w)$}
            \FOR{$\forall n \in V in\ random\ order$}
                \STATE \textcolor{commentcolor}{\textit{La communauté du n\oe{}ud devient la communauté majoritaire des voisins.}}\\
                \STATE $C(n) \leftarrow countmax( \{c(v),\ \forall v \in voisins(n)\}, w)$\\
            \ENDFOR
        \ENDWHILE
        \RETURN C
    \end{algorithmic}
\end{algorithm}

\paragraph{Extraction des plongements lexicaux.}

Une fois les communautés extraites, il reste à construire les plongements pour notre vocabulaire. Pour ce faire, nous considérons la distribution des liens de chaque n\oe{}ud à travers les communautés. 
Néanmoins, il s'agit de prendre en compte l'influence du degré du n\oe{}ud et de celui de ses voisins. Prenons un exemple pour clarifier : celui des mots \textit{escroc} et \textit{aigrefin}. 
Ces deux mots sont proches d'un point de vue sémantique. 
Par contre, \textit{escroc} est plus fréquent qu'\textit{aigrefin}. 
Il sera donc mécaniquement d'un degré pondéré plus élevé.
Une fois cette remarque faite, on se rend compte que si on considère seulement la distribution d'\textit{escroc} dans les communautés pour créer son plongement, la norme de son vecteur sera plus grande que celle d'\textit{aigrefin}.

Par ailleurs, la taille des communautés a une influence similaire. L'algorithme de détection de communautés aboutit (sauf exception) à une partition dont les tailles des communautés sont hétérogènes.
Si l'on ne tient pas compte de cet état de fait, les communautés les plus petites auront mécaniquement une composante plus faible que les grosses communautés dans les vecteurs.

Ainsi, si on note $e_n$ le plongement du n\oe{}ud représentant le mot $n$, $e_n \in {\rm I\!R}^{|C|}$ et $e_n^c$ la valeur de la composante correspondante à la communauté $c$ de $e_n$, cette valeur se calculera ainsi :
    \begin{minipage}{0.48\linewidth}
    \begin{equation}
        \hat{e}_n^c = \frac{1}{|N^c(e_n)|}\sum_{v \in N^c(e_n)}sppmi(n, v)
    \end{equation}
      
   \end{minipage}\hfill
   \begin{minipage}{0.48\linewidth}
        \begin{equation}
        e_n^c =  \frac{\hat{e}_n^c - \mu(\hat{e}_*^c)}{\sigma(\hat{e}_*^c)}
        \label{eq:zscore}
        \end{equation}
    \end{minipage}
    \label{eq:embedding}
Avec $N^c(e_n)=voisins(n)\cap C_c$, \textit{i.e.} l'ensemble des voisins du noeud représentant le mot $n$ appartenant à la communauté $c$, $\mu(\hat{e}_*^c)$ et $\sigma(\hat{e}_*^c)$ respectivement la moyenne et l'écart-type des valeurs de $\hat{e}_n^c$, $\forall n \in V$ et $sppmi$ une version normalisée de la $ppmi$ (Éq. \ref{eq:ppmi}) à valeur dans $\left[0, 1\right]$.

\begin{sidewaysfigure}
    \centering
    \includegraphics[width=0.6\textwidth]{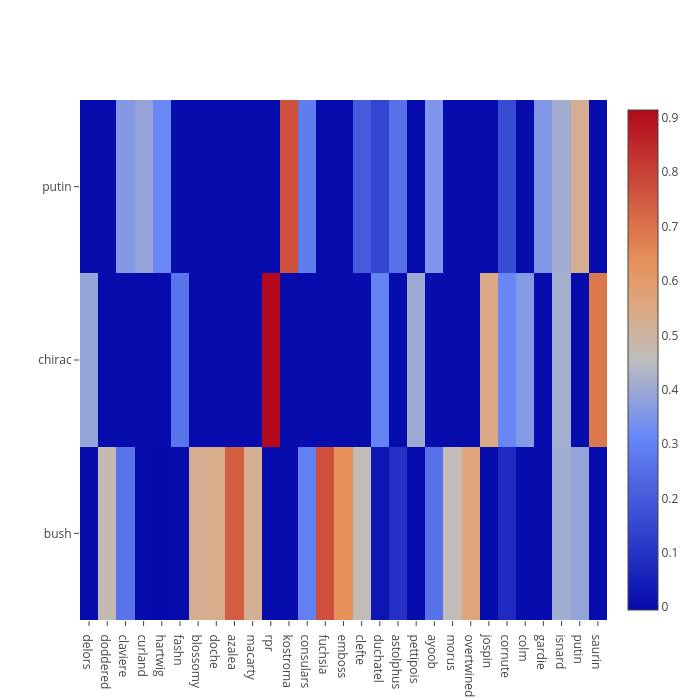}
    \caption{En abscisse, les étiquettes des 30 dimensions les plus caractéristiques des vecteurs de putin, chirac et bush (10 par vecteur). La couleur représente la valeur de la composante pour chacun des vecteurs.}
    \label{fig:chirac}
\end{sidewaysfigure}

L'utilisation de la $sppmi$ nous permet de contrebalancer l'influence du degré du n\oe{}ud et de celui de ses voisins, celle du \emph{z-score} (Éq.~\ref{eq:zscore})  l'influence de la taille des communautés. L'exemple de la Figure~\ref{fig:chirac} illustre le résultat une fois toutes les étapes réalisées, en proposant une visualisation des vecteurs de \emph{bush}, \emph{putin} et \emph{chirac} via les 30 dimensions les plus utiles pour la caractériser (10 par vecteur).

\section{Résultats}
\label{sec:resultats}
Nous débuterons cette section avec quelques évaluations empiriques purement qualitatives concernant la pertinence des dimensions exploitées (les communautés) et l'espace appris (les voisinages). Nous donnerons enfin des résultats quantitatifs 
 qui démontrent l'intérêt de l'approche. Sur notre corpus, après pré-traitement, nous aboutissons à une taille de vocabulaire qui est d'un peu plus de $135.000$ mots, ce qui correspond au nombre de n\oe{}uds du graphe. Après détection de communautés, nous obtenons environ $30.000$ communautés (voir la distribution de leurs tailles Figure~\ref{fig:distribution}), soit des vecteurs de taille $30.000$. En revanche, on constate qu'en moyenne, seulement 300 (environ) composantes du vecteur sont non-nulles, les vecteurs sont donc extrêmement creux.
 
\paragraph{Communautés et interprétabilité.}
En utilisant des méthodes d'étiquetage des communautés, il est possible d'évaluer empiriquement la pertinence de l'approche. Les communautés extraites constituent les dimensions des vecteurs qui semblent ainsi interprétables et cohérentes. 
Nous donnons ici en exemple trois communautés et leurs étiquettes caractéristiques :
\begin{itemize}
    \item ('officiel', 'republique', 'parisienne', 'couture', 'senat')
    \item ('copper', 'iron', 'stand', 'metal', 'upon')
    \item ('volleyball', 'handball', 'softball', 'badminton', 'basketball')
\end{itemize}
La première communauté regroupe les mots français du corpus.
La seconde concentre du vocabulaire lié aux métaux même si l'on peut constater qu'on y trouve des intrus (\emph{stand} et \emph{upon}).
Enfin, la dernière communauté regroupe du vocabulaire lié au sport. Ce sont les mêmes méthodes d'étiquetage qui sont utilisées dans la Figure~\ref{fig:chirac} pour étiqueter chaque dimension des vecteurs visualisés. On reconnaît un bon nombre de ces étiquettes comme par exemple \emph{Delors} (Jacques), \emph{rpr} (Rassemblement pour la république), \emph{Jospin} (Lionel) pour caractériser le vecteur de Jacques Chirac. Il est particulièrement intéressant de s'intéresser au vecteur \emph{bush}, où l'on trouve \emph{Mcarthy} (Joseph), ou encore \emph{Putin} (Vladimir) mais également des noms d'arbustes puisque c'est l'un des sens de \emph{bush} (\emph{azalea}, \emph{fuschia}). Grâce à l'interprétabilité du modèle, nous pouvons ainsi observer la façon dont celui-ci intègre la polysémie/l'homonymie.

\paragraph{Base canonique et interprétabilité.}
Avec une approche telle que celle de \emph{Word2Vec}, il est difficile d'interpréter les dimensions. Tout d'abord, supposer qu'il est possible d'interpréter les dimensions revient à faire l'hypothèse que chaque dimension peut être considérée indépendamment des autres, et que chaque dimension a un sens cohérent, i.e. qu'explorer les vecteurs colinéaires aux vecteurs de la base canonique de l'espace appris permettrait d'extraire ces \emph{sens}. Empiriquement, il est pourtant difficile d'affirmer cela. Considérons un modèle Word2vec à 300 dimensions appris sur le corpus Google News~\cite{mikolov2013distributed}, et prenons des contre-exemples simples. Soit $C$ la base canonique de l'espace de dimension $300$ de notre expérimentation telle que $C=\left\lbrace e_1=(1, 0, 0, \cdots, 0), e_2=(0, 1, 0, \cdots, 0), \cdots, e_{300}=(0, 0, 0, \cdots, 1) \right\rbrace$.
\begin{itemize}
    \item Considérons ainsi les 10 termes les plus proches de $e_4$ dans l'espace : Ginsburgs, Dinty Moore, jelly sandwiches, cheartier appetites, they'd, Fabens fliers, banana republics, isn, Chipotle burritos, payroll deduction.
    \item Pour $e_9$, nous obtenons les résultats suivants :
costliest natural disasters, counterparty defaults, mute button, closely scrutinized, Bernankes, damage Minsch, student Tyler Clementi, degraded Kenneth Merten, historian Bob Kreipke, Nishu Sood.
\end{itemize}
Il semble très difficile de tirer une quelconque cohérence dans les termes qui sont retournés, contrairement aux communautés précédemment citées. Les communautés sont en effet des objets concrets, des ensembles de mots du corpus qui sont particulièrement connectés ensemble, et elles peuvent de plus être considérées indépendamment les unes des autres.

Considérons maintenant notre modèle et les vecteurs canoniques de l'espace constitué via l'extraction des communautés. Dans notre modèle comme dans les autres, il est possible d'extraire les plus proches voisins d'un mot ou d'un vecteur en utilisant la similarité \emph{cosine} pour évaluer la distance entre deux vecteurs. Considérons dans les cas des deux exemples suivants les $10$ vecteurs les plus proches du vecteur canonique dont la composante non-nulle correspond à la communauté qui contient le mot \textbf{alcohol}, puis \textbf{petal} : 
\begin{itemize}
    \item mannite, dinitro, polyhydric, benzole, benzol, lactose, fermenter, disaccharide, bisulphide, reconverted.
    \item sepal, papilionaceous, floret, stamen, blotch, dewdrops, petals, bracts, corolla.

\end{itemize}
Dans le premier exemple, de manière générale, on obtient des termes liés à l'alcool directement : \emph{mannite} pour le manniotal qui est un alcool, \emph{polyhydric} parce que les alcools de sucre sont dits polyhydriques ; des termes liés au sucre qui est l'un des éléments de base pour la création d'alcool (\emph{lactose}, \emph{disaccharise}), au processus de création d'alcool (\emph{fermenter}), ou à la chimie (\emph{dinitro}, \emph{bisulphide}).
Dans le second exemple, tout ou presque est lié à la fleur : les sépales (\emph{sepal}), les étamines (\emph{stamen}), \emph{papilionaceous} qui est une fleur, \emph{drewdrop} qui signifie goutte de rosée, \emph{bract} qui est une petite feuille, \emph{corolla} qui est un synonyme de pétale. Dans ces deux cas, il existe un fort recouvrement entre les $10$ plus proches voisins mentionnés ci-dessus et les représentants les plus caractéristiques de la communauté (étiquettes).

Un autre exemple parlant est celui du vecteur $e_{746}$ de notre modèle dont les $10$ plus proches voisins sont :
sifteen, fiftyfour, fortyseven, fiftyseven, sixtyfive, fortysix, fiftyfive, sixtyseven, twentyeight, twentyseven.

La distance \emph{cosine} peut comme dans les autres modèles être exploitée pour étudier la similarité entre les termes du vocabulaire, pas seulement avec les vecteurs canoniques. Ainsi, la liste de voisins suivante fournit quelques exemples de résultats illustrant le bon fonctionnement de la méthode :

\begin{itemize}
    \item metal: (metals, metallic, iron, copper, steel, alloy, aluminium, oxides, chromium)
    \item picture: (pictures, portrait, image, painting, view, images, depiction, portrayal, painted)
    \item salad: (mayonnaise, ketchup, lettuce, tomato, vegetables, sauce, celery, mashed, cheese)
    \item mars: (altimeter, orbiter, venus, saturn, jupiter, orbit, pioneer, planets, planet)
    \item news: (television, cnn, bbc, pathe, nbc, tidings, newspapers, cbs, gaumont)
\end{itemize}

\paragraph{Comparaison à l'état de l'art.}
Pour obtenir des résultats quantitatifs, nous comparons notre approche à celles de l'état de l'art en considérant deux tâches d'évaluation~\cite{schnabel_evaluation_2015} :
\begin{description}
\item[Similarité] La tâche de similarité se présente comme une base de données de paires de mots, avec pour chaque paire un score associé.
Le score de similarité entre deux mots est issu d'une évaluation humaine. 
La qualité du modèle peut donc être évaluée en calculant la corrélation entre le vecteur de score humain et le vecteur de distances entre les vecteurs appris.
Une corrélation linéaire (coefficient de \emph{Spearman} proche de $1$) correspond à un modèle complètement en accord avec l’évaluation humaine.
\item[Catégorisation] La tâche de catégorisation se présente comme une base de données de paires (mot, catégorie). 
Le but est de réussir à regrouper des mots en différentes catégories en utilisant les vecteurs appris. Pour faire cela, on opère une analyse de regroupement sur les vecteurs appris. On évalue ensuite le modèle en calculant la pureté entre les regroupements et la catégorisation humaine.
\end{description}

\begin{table}[h]
\centering
\resizebox{\textwidth}{!}{%
\scriptsize
\begin{tabular}{c c c c}
\hline \hline Benchmark & Notre modèle & État de l'art dim=300 & État de l'art dim=50\\ \hline \hline
\multicolumn{4}{c}{Similarité} \\ \hline
MEN & 0.650 & 0.809 & 0.720\\
SimLex & 0.364 & 0.427 & 0.309\\
RG65 & \textbf{0.803} & 0.790 & 0.763\\ \hline
\multicolumn{4}{c}{Catégorisation} \\ \hline
ESSLI1a & 0.75 & 0.818 & 0.773\\
ESSLI2b & \textbf{0.775} & 0.750 & 0.775\\
ESSLI2c & 0.6 & 0.667 & 0.556\\
\end{tabular}
}
\caption{\label{tab:res}Résultats sur les tâches de catégorisation et de similarité en comparaison de l'état de l'art.}
\end{table}

Nous utilisons la librairie \emph{word-embeddings-benchmarks}\footnote{https://github.com/kudkudak/word-embeddings-benchmarks} pour réaliser nos évaluations~\cite{jastrzebski2017evaluate}. Nous comparons nos résultats à ceux obtenus avec des plongements pré-entraînés accessibles en ligne en utilisant cette librairie. Les plongements utilisés sont ceux obtenus via les méthodes Glove~\cite{pennington_glove:_2014}, NMT~\cite{hill2014embedding}, HDC et PDC~\cite{sun2015learning}, Skip-gram~\cite{mikolov2013distributed} et Lexvec~\cite{salle2016enhancing}.

Les résultats Table~\ref{tab:res} sont encourageants, ils montrent que notre approche est pertinente. Pour chaque tableau, nous comparons les résultats de notre méthode aux meilleurs résultats des méthodes de l'état de l'art citées, pour $50$ et $300$ dimensions. Sur deux corpus (en gras dans la Table), l'un exploité pour la tâche de similarité, l'autre pour la tâche de catégorisation, notre méthode obtient des résultats comparables à celles de l'état de l'art auxquelles nous nous comparons. Dans le reste des cas, notre méthode obtient des résultats supérieurs aux performances des approches de l'état de l'art paramétrés pour retourner des vecteurs en dimension $50$, mais inférieurs lorsque ces vecteurs sont en dimension $300$.
En accord avec l'état de l'art nos meilleurs résultats sont obtenus pour les plus grandes tailles de fenêtre ($f=5$ dans notre cas).

\section{Discussion et perspectives}
\label{sec:discussion}

Nous avons décrit une méthode originale d'apprentissage de plongements lexicaux basée sur une approche réseaux complexes. Nous proposons d'utiliser les communautés détectées sur le réseau de co-occurrences représentant le corpus comme dimensions de nos plongements. Les vecteurs sont ensuite directement extraits de la distribution des liens de chaque n\oe{}ud à travers la structure communautaire. Les résultats qualitatifs et quantitatifs montrent la pertinence de l'approche qui obtient des scores comparables à l'état de l'art. Néanmoins, une étude avec les mêmes méta-paramètres (corpus, taille de fenêtre) semblent nécessaire pour se situer exactement par rapport à l'état de l'art.

Cette approche a pour avantage de fournir des dimensions qui sont des objets concrets, physiquement existants : les communautés. Ces dimensions semblent donc interprétables : il est possible de consulter le contenu de ces communautés, de les étiqueter avec des éléments caractéristiques. Néanmoins, cela ne garantit pas l'interprétabilité des vecteurs appris. Pour que ces vecteurs soient interprétables, il s'agit à notre sens de réunir deux conditions. La première, est de disposer d'un étiquetage suffisamment précis pour qu'il soit tout à fait compréhensible. La seconde nécessite d'avoir un vecteur de taille raisonnable, ou du moins un vecteur creux afin de ne pas avoir trop de communautés à inspecter.
Ces questions sont en lien direct avec le paramétrage des algorithmes de détection de communautés et constituent des perspectives directes de notre travail. Nous souhaitons en effet travailler à évaluer l'interprétabilité des vecteurs extraits par notre méthode par des humains.

De plus, notre méthode permet l'extraction rapide du plongement d'un mot ou d'une expression. Le calcul de ce vecteur découle en effet directement de la connectivité du n\oe{}ud qui représente le mot, de la façon dont ses liens se dispersent au sein de la structure communautaire. Ainsi, le calcul du vecteur d'un nouveau mot ou d'une expression composée ne nécessite pas de réapprendre un modèle, mais simplement d'ajouter le terme au réseau pour extraire le vecteur.

Enfin, les langues évoluent avec le temps : le sens des mots change ou de nouveaux sens apparaissent. Ces évolutions de la langue ont été décrites, notamment par~\citealt{bloomfield1983introduction,mitra2015automatic}. Des travaux considèrent des méthodes automatiques basées sur les plongements lexicaux et de grands corpus temporels pour la détection de ces néologismes sémantiques~\cite{tang_state---art_2018}. Ces méthodes peuvent être séparées en deux classes. La première classe est celle des méthodes \emph{diachroniques} : elles discrétisent le temps et séparent ainsi le corpus en plusieurs sous-corpus. Sur chacun de ces sous-corpus, les auteurs proposent d'apprendre des plongements lexicaux puis d'aligner les espaces appris entre les sous-corpus deux à deux~\cite{hamilton_diachronic_2016}. Ces approches sont basées sur l'hypothèse très forte qu'il est possible d'aligner des espaces différents issus d'algorithmes non déterministes aboutissant à des résultats sous-optimaux. La seconde classe, celle des méthodes \emph{dynamiques}, propose une optimisation globale de tous les plongements du vocabulaire à travers le temps, aboutissant à un problème gourmand en calcul et très difficile~\cite{bamler_dynamic_2017}. 
Notre approche peut permettre d'ouvrir le champ à de nouveaux travaux basés sur les algorithmes de détection de communautés incrémentaux~\cite{xie2013labelrankt}. Cela permettrait ainsi de s'abstraire d'une optimisation globale coûteuse, et de contourner l'hypothèse d'alignement diachronique des espaces.

\bibliographystyle{taln2019}
\bibliography{biblio}

\end{document}